\documentclass[11pt]{article}

\usepackage{microtype} 
\usepackage{booktabs}  
\usepackage{url}  

\usepackage{amsmath}
\usepackage{amsthm}
\usepackage{graphicx}

\usepackage[final]{mais}
\usepackage{natbib}

\title{Physical Simulation for Multi-agent Multi-machine Tending}

\author[1]{\nameemail{Abdalwhab Abdalwhab}{Abdalwhab.Bakheet@gmail.com}}

\author[2]{\nameemail{Giovanni Beltrame}{giovanni.beltrame@polymtl.ca}}

\author[1]{\nameemail{David St-Onge}{David.st-onge@etsmtl.ca}}

\affil[1]{Department of Mechanical Engineering, ETS Montreal}

\affil[2]{Department of Computer and Software Engineering, Polytechnique Montréal}

\hypersetup{%
  pdfauthor={Abdalwhab Bakheet Mohamed Abdalwhab}, 
  pdftitle={MachineTendingMAIS2024},
  pdfsubject={},
  pdfkeywords={Reinforcement Learning, Multi-agent, Multi-machine Tending, Mobile robots }
}

\begin{document}

\maketitle
The manufacturing sector like many other sectors was recently affected by workforce shortages, a problem that automation and robotics can heavily minimize \cite{Logan2022laborshortage}. Simultaneously, Reinforcement learning (RL) offers a promising solution where robots can learn to perform tasks through interaction and feedback from the environment \cite{singh2022reinforcement}. However, despite their success in numerous simulation environments, we still don't see many real-world deployments of RL robotic solutions. In fact, many researchers either oversimplify the targeted real-world scenario such as \cite{wu2023daydreamer} or do not even evaluate their model in physical robots \cite{lu2022socially,na2022bio}.

It is known that training RL policies directly in real robots can be expensive, time-consuming, labor-intensive, and maybe even dangerous, that is why it makes sense to try to leverage training in simulation. At the same time, the gap between the simulation environments that typical RL policies are trained on and the real world is significant. This leads to a huge loss in performance during deployment, some works attempt to shorten this gap with techniques like domain randomization and adaptation, knowledge
distillation, meta-learning, and imitation learning \cite{zhao2020sim}. Another direction is improving the real-world representation in the simulator such as Meta Habitat \cite{puig2023habitat3} and Isaac Lab \cite{mittal2023orbit}.

Differently, in this work, we leveraged a simplistic robotic system to work with "real" data without having to deploy large expensive robots in a manufacturing setting. The aim is to study the challenging scenario of multi-agent multi-machine tending manufacturing setups. In this setup, a fleet of mobile manipulators navigate between production machines to feed raw materials pick up ready parts, and drop them in the designated storage area.

We started by designing a multi-agent multi-machine tending scenario in VMAS \cite{bettini2022vmas} a simulator specifically designed for multi-agent reinforcement learning (MARL) research. We then used the scenario to train the well-established MARL model: MAPPO \cite{yu2021surprising} to get a model that works well in simulation. After that, to further study the model with hardware-in-the-loop, we choose simple custom-made table-top robots adapted from the original Stanford's Zooids \cite{le2016zooids}.

\begin{figure}[ht]
  \centering
  \includegraphics[width=0.7\linewidth]{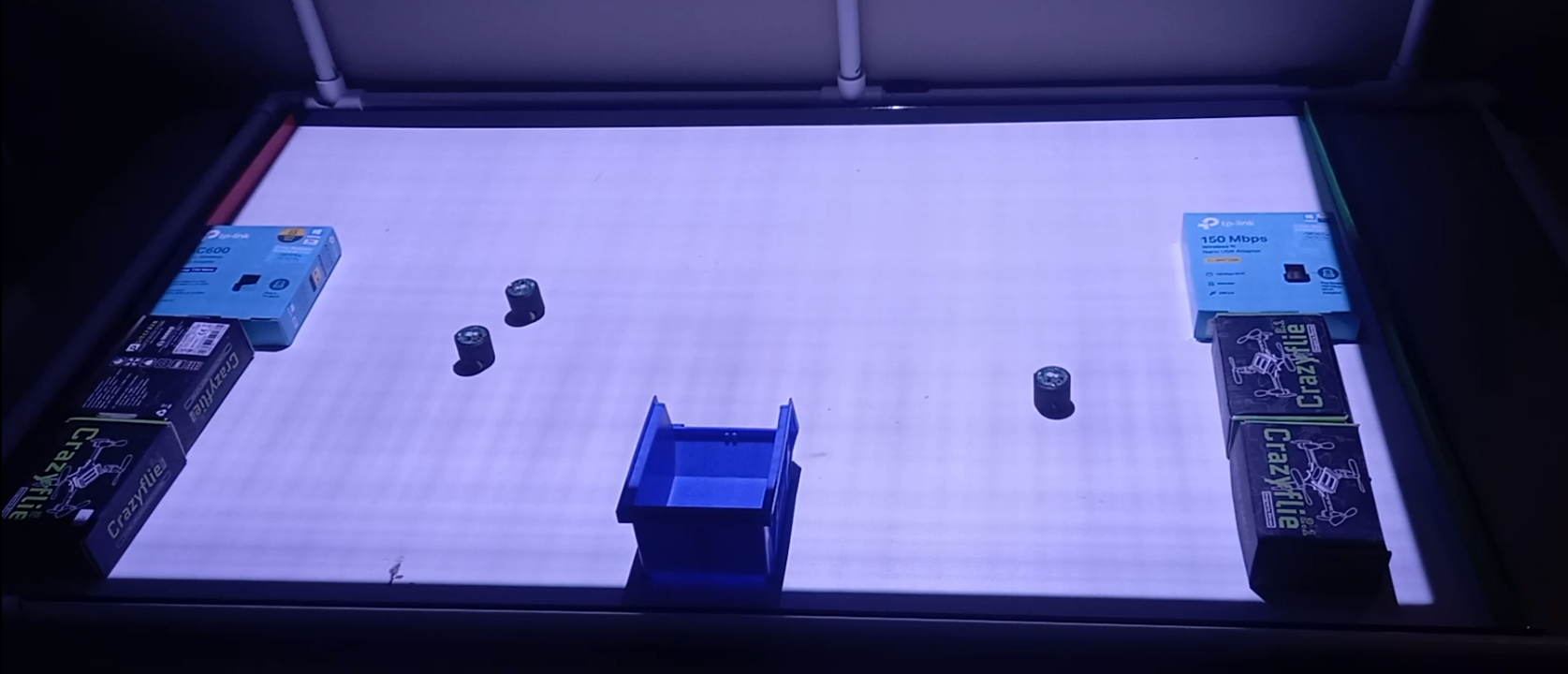}
  \caption{The real-world experiment arena: 3 Zooids, storage area (blue in the middle), machines (blue boxes), and machines' blockers (black)}
  \label{fig:RealArena}
\end{figure}

In the experiment setup, we designed a real-world table-top arena shown in figure \ref{fig:RealArena} that mimics the scenario we designed in simulation with three Zooids scattered in the middle, a small blue shelf down in the middle representing the storage area, blue boxes on the two sides representing the machines, and black boxes representing machines' blockers. 

In this setup, the simulation is always running in a ground station, and the robots in the arena try to achieve the task by mimicking the simulation. Each robot in the arena is represented by one agent in the simulation, and the position of that agent is sent for the robot in the arena to follow. Despite the difference in dynamics (omnidirectional agents in simulation, compared to differential motion Zooids and smaller size machines and storage area) the robots were able to depict the same behavior as in the simulation, going to the machines to get the ready parts and then going to the storage area to deliver them.

Those experiments allow us to acknowledge a couple of deployment challenges, namely the robot's onboard computational power, localization, and communication. Since the Zooids have small computational power everything was running on a central ground station. Moreover, Zooids localize themselves by detecting gray-coded patterns projected from the top by a projector, but for an actual deployment a more robust localization setup is needed such as combining wheel odometry, the IMU measurements, and UWB localization \cite{imran2024lab}. Furthermore, we noticed a communication bottleneck when trying to get the Zooids' actual positions back to the ground station through a central Arduino-based antenna which can be eliminated with direct inter-robot communication, with each robot sharing its location with its neighbors.

\begin{figure}[ht]
  \centering
  \includegraphics[width=0.31\linewidth]{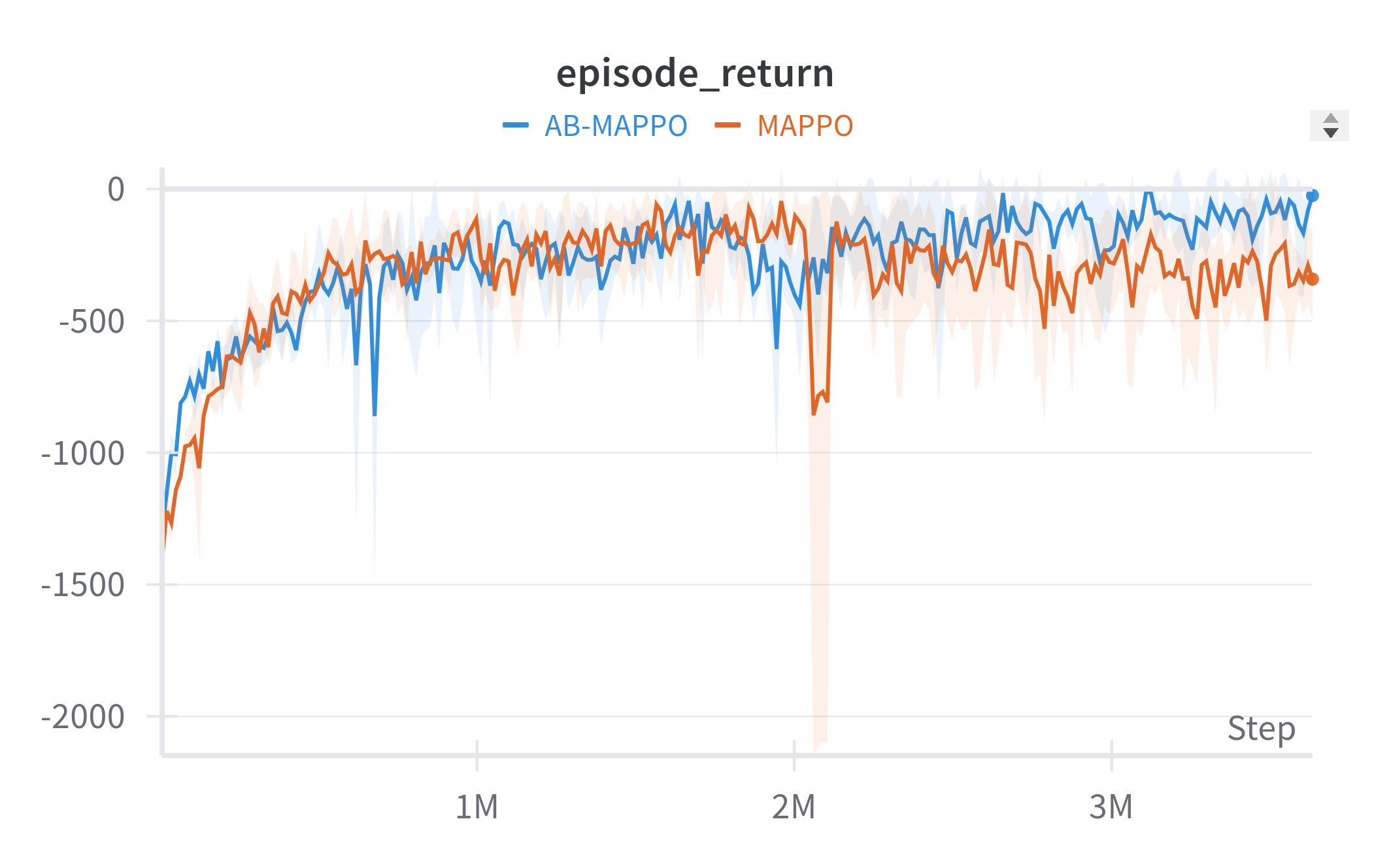}
  \includegraphics[width=0.31\linewidth]{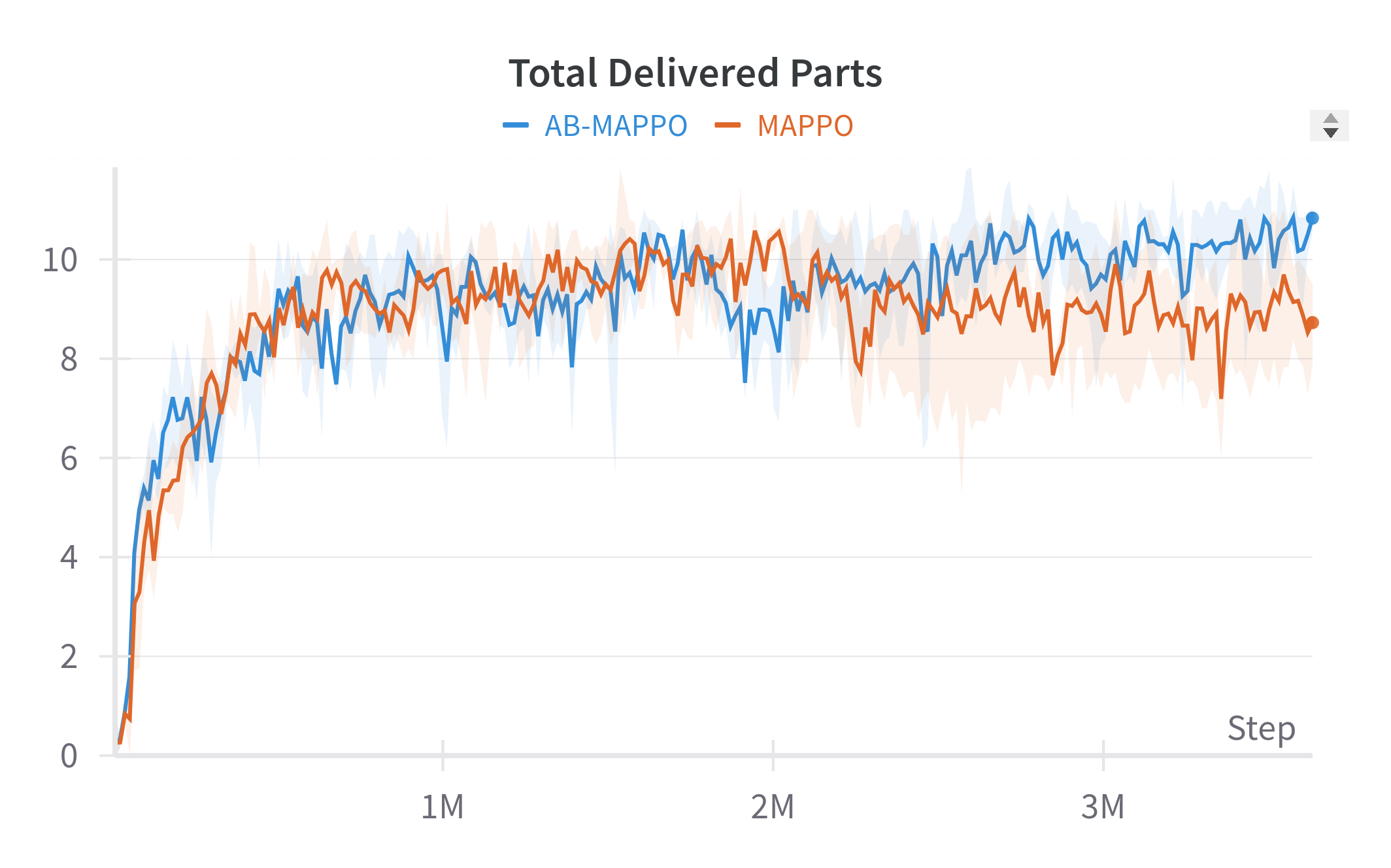}
  \includegraphics[width=0.31\linewidth]{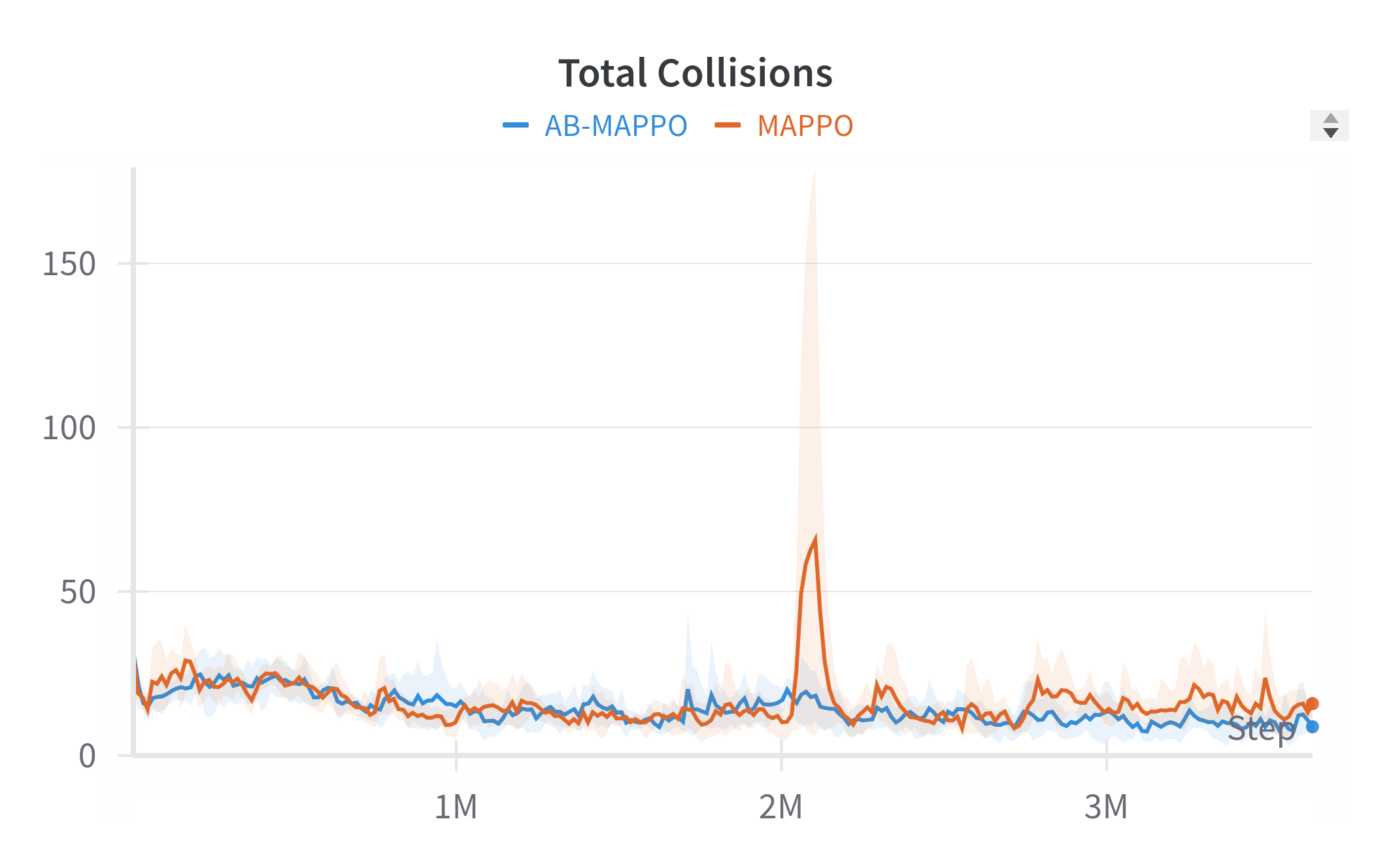}
  \caption{From the left: the episode total return, total delivered parts, and total collisions for AB-MAPPO (blue) compared to MAPPO (red)}
  \label{fig:episode_return}
\end{figure}

In summary, our experiments with the Zooids provided an initial understanding of the real deployment challenges, and we plan to experiment with more realistic setups using more robust robotic platforms and leveraging our improved model AB-MAPPO (Attention-based encoding mechanism for MAPPO) \cite{abdalwhab2024learning}. Figure \ref{fig:episode_return} depicts the episode total return, followed by total delivered parts and total collisions of AB-MAPPO compared to MAPPO. It is worth mentioning that AB-MAPPO not only collects more rewards, but also performs better in terms of parts delivery, and number of collisions.

\begin{acknowledgements}

We thank NSERC USRA and Discovery programs for their financial support. We also acknowledge the support provided by Calcul Québec, Compute Canada.

\end{acknowledgements}


\bibliography{references}
\bibliographystyle{mais}



\appendix

\end{document}